%% file: main.tex
\documentclass[10pt,twocolumn,letterpaper]{article}

\usepackage[pagenumbers]{cvpr}


\usepackage{multirow}
\usepackage{booktabs}

\definecolor{cvprblue}{rgb}{0.21,0.49,0.74}
\usepackage[breaklinks,colorlinks,allcolors=cvprblue]{hyperref}

\renewcommand{\paragraph}[1]{\vspace{.1em}\noindent\textbf{#1}}

\title{FineDex: Bridging Fine-Grained Human--Object Interaction Generation to Dexterous Humanoid Manipulation}

\author{
    Yujia Zeng$^{1}$ \quad
    Yuxin Chen$^{1}$ \quad
    Yuzhen Song$^{1}$ \quad
    Runhan Huang$^{2}$ \\[3pt]
    Chensheng Peng$^{1}$ \quad
    Yiyang Shao$^{1}$ \quad
    Koushil Sreenath$^{1}$ \\[3pt]
    Masayoshi Tomizuka$^{1}$ \quad
    Yilun Du$^{2}$ \\[6pt]
    $^{1}$University of California, Berkeley \qquad
    $^{2}$Harvard University
    \\
    {\small \url{https://finedex.github.io/}}
}

\makeatletter
\apptocmd{\@maketitle}{%
    \begin{center}
        \includegraphics[width=\textwidth]{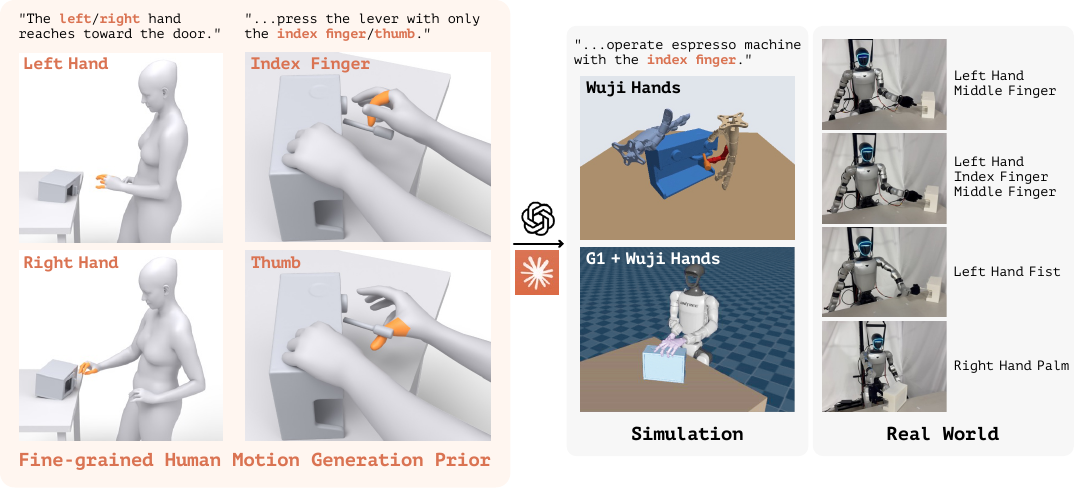}
        \captionof{figure}{
            \textbf{Fine-grained instruction-conditioned dexterous manipulation.}
            FineDex generates human--object interaction priors that capture
            language-specified hand selection, finger-level contacts,
            and articulated-object interactions.
            These priors guide agentic dexterous manipulation,
            enabling precise instruction following in simulation
            and on real humanoid robots.
        }
        \label{fig:teaser}
    \end{center}
     \vspace{0.8em}
}{}{}
\makeatother

\begin{document}
\maketitle

\input{sec/0_abstract}
\input{sec/1_intro}

\input{sec/2_relatedwork}

\input{sec/3_method}
\input{sec/4_experiment}
\input{sec/5_conclusion}

{
    \small
    \bibliographystyle{ieeenat_fullname}
    \bibliography{main}
}


\end{document}

%% file: sec/0_abstract.tex
\begin{abstract}

Dexterity is, at its core, finger-level control, yet existing dexterous manipulation and
human--object interaction (HOI) generation methods rarely let one specify how an
interaction is performed: five-fingered hands are often reduced to grippers, or finger usage is
copied from individual human demonstrations. We present FineDex, a framework that makes finger-level intent both
expressible and executable. FineDex learns coordinated human body, hand, and object motions that
follow fine-grained instructions on hand selection, individual fingers, and contacts with
articulated objects. 
To bridge generation and execution, FineDex leverages generated interactions
as actionable priors for an agent that discovers physically feasible
contact strategies through practice in a physics-based digital twin.
By adapting at test time without per-instance retraining,
FineDex generalizes
across fine-grained instructions and object configurations.
Experiments demonstrate state-of-the-art HOI generation quality 
and fine-grained controllability, substantial improvements in dexterous
manipulation success, and successful
deployment on a real dexterous humanoid robot.
\end{abstract}

%% file: sec/1_intro.tex
\section{Introduction}
\label{sec:intro}

Dexterity is, at its core, finger-level control: humans press with a fingertip,
hook with a specific finger, brace with the palm, and switch hands when one is busy.
Yet existing dexterous manipulation and human--object interaction (HOI)
generation methods are evaluated mainly by task completion
alone: many effectively use a five-fingered hand as a gripper for
pick-and-place~\cite{DexGraspNet,UniDexGrasp}, 
while others copy whatever finger usage appears in a specific human
demonstration~\cite{REGRIND,DexCap}. Meanwhile, human interaction data offers a scalable source of rich
finger-level behaviors. The missing link is a model that can \emph{express} such intent
on demand---which hand, which finger, which contact---and a way to \emph{realize}
it on a physical robot.

However, existing methods fall short on both fronts. On the expression side, text-conditioned HOI generators
synthesize coordinated human--object motion~\cite{Text2HOI,HOIDiNi,MADHOI,CoDA},
but they respond almost only to task-level descriptions: asked to use the left hand or a
particular finger, they produce the same interaction regardless. On the realization side,
generated motions are not physically exact, and differences
in robot kinematics and contact dynamics make direct
motion replay unreliable. Existing pipelines bridge this gap by training
reference-tracking policies~\cite{REGRIND,EmbodyHOI},
but their reliance on reference-specific training
limits scalability to diverse fine-grained instructions,
object configurations, and robot-specific constraints.

Our key insight is to decouple fine-grained intent generation
from its physical realization, treating generated HOI
as a sketch of the intended interaction.
\emph{What a human would do} is best captured by a generative prior
learned from fine-grained human interaction data.
\emph{How the robot can do it} is best discovered through
test-time practice in a physics-based digital twin.
Generation supplies the finger-level intent;
practice supplies its physical grounding.

Building on this observation, we introduce \textbf{FineDex}, a framework that makes
finger-level intent both expressible and executable. As shown in Fig.~\ref{fig:teaser}, FineDex learns
to generate coordinated body, hand, and object motions
conditioned on fine-grained instructions specifying hand selection,
individual finger motions, and designated contacts. For execution, FineDex treats generated interactions as
actionable priors for an agent that proposes contact programs, practices them under
randomized dynamics, and commits to the most reliable behavior, which
is then executed on the robot. Because adaptation happens through practice, the same system
generalizes to new instructions, robot-specific constraints,
and object configurations without retraining.

To measure what task success alone cannot, we introduce FingerBench, which evaluates
whether generated motions follow hand- and finger-level instructions and whether robots
realize them in physics-based execution. Experiments demonstrate state-of-the-art HOI generation quality
and finger-level controllability. In physics-based execution, practice turns generated priors into reliable
contact behaviors where direct replay rarely succeeds while preserving designated finger contacts. On a real Unitree G1 humanoid, FineDex completes
articulated-object tasks under constraints it was never trained for, such as operating
with the left hand only or with a single designated finger.

\noindent Our main contributions are summarized as follows:

\begin{itemize}
    \item \textbf{Finger-level HOI generation.} A generator of coordinated body, hand,
    and articulated-object motion that follows fine-grained instructions on hand selection,
    specific fingers, and designated contacts, with state-of-the-art controllability
    and quality.
    \item \textbf{Practice-grounded dexterous execution.} An agent that grounds generated
    interactions through practice, adapting to fine-grained instructions and object configurations
without retraining, with deployment on a real humanoid.
    \item \textbf{FingerBench and fine-grained ARCTIC annotations}:
    a benchmark for evaluating fine-grained instruction following in HOI generation
    and robotic execution, and fine-grained annotations for
    training instruction-conditioned HOI models, with the latter to be publicly released.
\end{itemize}

%% file: sec/2_relatedwork.tex
\begin{figure*}[!t]
    \centering
    \includegraphics[width=\textwidth]{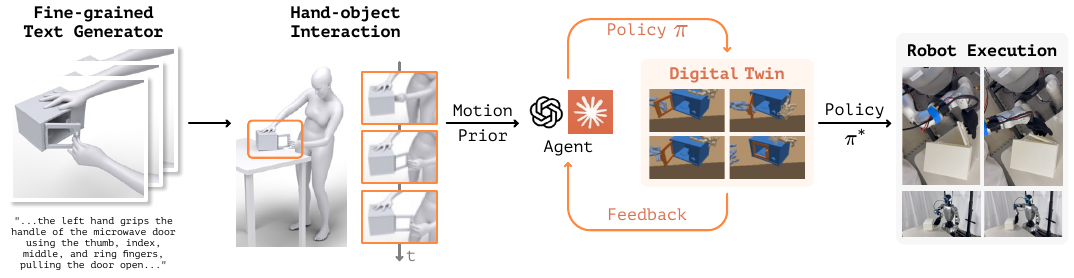}
    \caption{
        \textbf{Overview of FineDex.}
        A language-conditioned generator synthesizes
        fine-grained human--object interactions.
        The generated interaction $\mathbf{x}$ serves as a prior
        for an agent that proposes contact programs $\pi$ and
        refines them through feedback from a physics-based digital
        twin. The most reliable program, $\pi^{\star}$, is then
        executed on a dexterous humanoid robot.
    }
    \label{fig:overview}
\end{figure*}

\section{Related Work}

\subsection{Human--Object Interaction Generation}
Text-conditioned HOI generation synthesizes coordinated hand, body, and object motion. Hand-centric methods generate contact-aware interactions through
diffusion~\cite{Text2HOI,DiffH2O} or masked autoregressive
diffusion~\cite{MADHOI}, while whole-body methods generate human--object
motion~\cite{OMOMO,CHOIS,CoDA}. HOIDiNi~\cite{HOIDiNi} improves contact realism through
diffusion noise optimization, F-HOI~\cite{F-HOI} aligns fine-grained interaction states
with language, and physics-based approaches synthesize dexterous grasping and articulated
interactions~\cite{ArtiGrasp,GraspXL,HOIFHLI}. Despite improved semantic and contact fidelity,
instruction-level control over hand selection and
individual finger contacts remains underexplored.
Our work addresses this gap, evaluated through FingerBench.

\subsection{Learning Dexterity from Human Data}
Human demonstrations provide a rich source of dexterous behavior. Retargeting methods
transfer captured or video-derived hand motion to robot
hands~\cite{DexMV,DexCap,ManipTrans,DexMachina,SPIDER}, often refined by residual
reinforcement learning~\cite{REGRIND,ManipTrans}, and DexMimicGen~\cite{DexMimicGen}
scales demonstrations through trajectory generation. At the humanoid level, motion
trackers reproduce large libraries of reference behaviors~\cite{SuperMimic}, and
RoboSTAR~\cite{RoboSTAR} transfers generated sign-language motion, including fine finger
articulation, to humanoid hands without object contact. EmbodyHOI~\cite{EmbodyHOI}
adds embodiment-aware kinematic guidance during HOI generation, while
FlowHOI~\cite{FlowHOI} executes contact-aware generated HOI trajectories on robot arms.
 These methods primarily inherit finger usage from reference
demonstrations, with physical execution often relying on
imitation or reference-tracking policies.
FineDex instead enables instruction-specified finger control,
grounded through test-time practice without per-reference training.

\subsection{Language-Guided Agentic Manipulation}
Language-model agents plan robot behavior through affordance-grounded skill
selection~\cite{SayCan}, code generation~\cite{CodeAsPolicies}, and spatial or keypoint
constraints~\cite{VoxPoser,ReKep}. Simulation-driven approaches synthesize rewards and
tasks for policy learning~\cite{Eureka,RoboGen}, and RAX~\cite{RAX} acquires
manipulation skills through online exploration with executable robot programs.
FineDex connects agentic planning with generative HOI priors.

%% file: sec/3_method.tex
\section{Method}
\label{sec:method}

As shown in Fig.~\ref{fig:overview}, FineDex decomposes finger-level manipulation into two stages. Given an articulated object $\mathcal{O}$ and an
instruction $\ell$ that may specify the acting hand, finger motions, or
designated contacts, a generator expresses \emph{what a human would do} as a
human--object interaction $\mathbf{x}$ (Sec.~\ref{sec:method_generator}).
An agent then determines \emph{how the robot can realize it} by proposing and refining contact programs through practice (Sec.~\ref{sec:method_agent}).

\subsection{Fine-Grained HOI Generation}
\label{sec:method_generator}

\begin{figure*}[t]
    \centering
    \includegraphics[width=\textwidth]{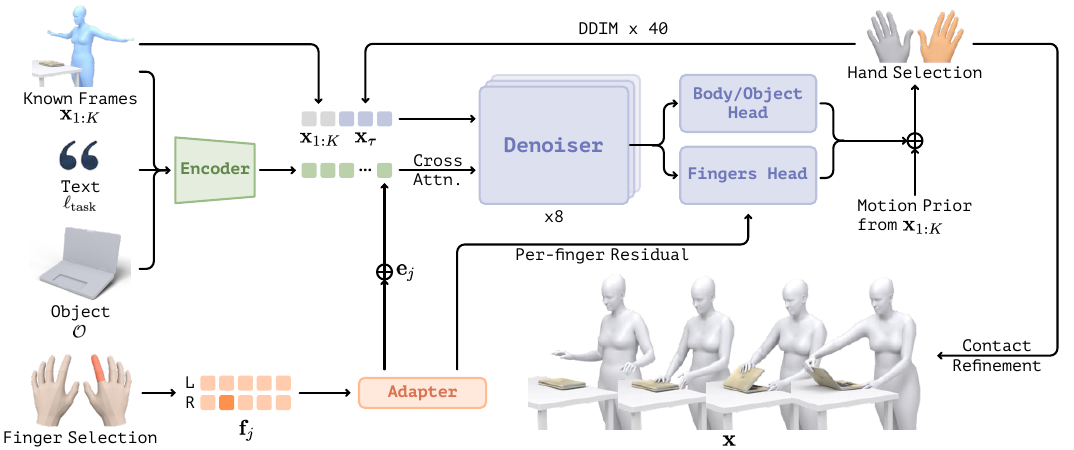}
    \caption{
        \textbf{Architecture of the FineDex generator.}
        The generator supports all articulated objects in ARCTIC,
        integrating language conditioning, articulated object geometry,
        and anatomical hand features within a diffusion-based framework.
        A finger-specific adapter injects control signals into
        anatomical anchor tokens while the pretrained backbone
        stays frozen.
        At inference, dual-trajectory sampling edits the commanded
        fingers while preserving the rest of the motion.
    }
    \label{fig:generator}
\end{figure*}

As shown in Fig.~\ref{fig:generator}, our generator combines
language, articulated-object geometry, and anatomical hand features to
synthesize coordinated human body, hand, and object motion. We train it
on ARCTIC~\cite{ARCTIC} with our fine-grained language annotations,
which describe hand roles, finger actions, contact regions, and object-state
changes.

\paragraph{Motion representation.}
We represent an interaction as $\mathbf{x}=\{\mathbf{x}^{t}\}_{t=1}^{N}$,
where each frame contains the human root pose, body joint rotations,
bilateral hand articulations, and object state.
The object state comprises a rigid base transform
$\mathbf{T}^{\mathcal{O}}_t\in SE(3)$ and an articulation angle
$\theta_t$, preserving hinge-constrained object motion.

\paragraph{Geometry-conditioned diffusion.}
We learn a conditional diffusion model
$p_{\phi}(\mathbf{x}\mid\ell,\mathcal{O},\mathbf{x}^{1:K})$,
where $\mathbf{x}^{1:K}$ are known context frames. Its conditioning
memory consists of token-level language features, point features of
the sampled object surface, an object-state token encoding the relative
base pose and hinge configuration, and anatomical anchor tokens
distributed across the fingers. A Transformer denoiser uses frame
queries to attend to these features. At diffusion step $s$, it predicts the clean sequence as a
residual to a damped constant-velocity extrapolation of the root, body,
and object channels of the context, with finger channels holding the
last context pose:
\begin{equation}
    \widehat{\mathbf{x}}_0=
    \mathcal{P}(\mathbf{x}^{1:K})+
    D_{\phi}(\mathbf{x}_s,s;\mathcal{C}),
    \label{eq:hoi_denoise}
\end{equation}
where $\mathcal{C}$ denotes the conditioning memory. With a single
context frame, $\mathcal{P}$ reduces to repeating it. Overlapping,
context-conditioned windows extend the model to longer interactions.
We train with reconstruction, kinematic, and contact-weighted losses,
and sample with DDIM and classifier-free guidance on text, keeping the
object and structural controls fixed across guidance branches.

\paragraph{Hand selection.}
We separate the task description from structured hand and finger
commands. To control the acting hand, we guide each denoising step
with low-dimensional, temporally interpolated body-pose offsets.
The guidance penalizes displacement of the stationary hand's anchor
centroid and wrist, while attracting the acting hand toward an
object- and part-dependent contact prototype estimated from training
interactions. The offsets are optimized at sparse temporal knots and
interpolated across the motion window. The inactive arm joints are
additionally fixed to their observed configuration.

\paragraph{Compositional finger motion control.}
We introduce a lightweight adapter for finger-specific motion
control without retraining the diffusion backbone.
Each control token encodes a finger identity and keyframe curl targets.
Trained on synthetic finger edits, the adapter injects
control residuals into the corresponding anatomical anchor tokens
and predicts finger-specific joint corrections.
Unlike direct joint interpolation, the learned edits account
for the surrounding hand--object motion, reducing implausible
hand poses and jitter. 

At inference, two denoising trajectories share the same initial noise:
a reference branch with task-level text and hand-selection guidance,
and a branch with finger-control conditioning. We combine their
clean-motion predictions at each step:
\begin{equation}
    \widehat{\mathbf{x}}_0^{\mathrm{edit}}=
    \widehat{\mathbf{x}}_0^{\mathrm{ref}}+
    \mathbf{M}_{\mathcal{F}}\odot
    \left(\widehat{\mathbf{x}}_0^{\mathrm{ctrl}}-
    \widehat{\mathbf{x}}_0^{\mathrm{ref}}\right),
    \label{eq:hoi_finger_edit}
\end{equation}
where $\mathbf{M}_{\mathcal{F}}$ selects the commanded finger
channels. This decoupled sampling preserves unconstrained body,
object, and hand motion while editing the specified fingers.

\paragraph{Finger-specific contact control.}
Finger articulation and finger contact impose distinct constraints.
Given a designated contacting set $\mathcal{F}^{\star}$, we refine
the generated motion by optimizing temporal offsets on the acting
arm and fingers:
\begin{equation}
    \min_{\boldsymbol{\Delta}}\quad
    \mathcal{L}_{\mathrm{touch}}+
    \mathcal{L}_{\mathrm{pen}}+
    \lambda_e\mathcal{L}_{\mathrm{excl}}+
    \lambda_a\mathcal{L}_{\mathrm{smooth}}.
    \label{eq:hoi_finger_contact}
\end{equation}
The touch term attracts sampled distal pads of the designated fingers
to the moving-part surface during contact frames, using the articulated-part geometry. The penetration term
keeps the acting hand's finger segments outside the moving part; for
exclusive-finger instructions, $\mathcal{L}_{\mathrm{excl}}$
additionally keeps non-designated fingers and the palm away from it; and
$\mathcal{L}_{\mathrm{smooth}}$ limits additional wrist and fingertip
acceleration. The refinement is applied post hoc to the sampled motion,
keeping the other hand, the torso, and the object fixed.
This allows contacts that never occur in the training data, such as
touching a part with a finger the human did not use.

\subsection{Generation-Guided Agentic Manipulation}
\label{sec:method_agent}

To bridge HOI generation and dexterous robot manipulation,
FineDex treats generated interactions as actionable priors
rather than trajectories to replay.
The generated HOI provides hand roles, approach directions,
contact configurations, and target object articulation,
but may not satisfy the robot's kinematic and physical constraints.
We therefore separate instruction constraints from motion guidance:
designated hands and fingers restrict admissible contact programs,
while the generated HOI guides contact selection and
support-hand placement.
An agent then adapts contact configurations and execution
parameters through test-time practice, without per-instance training.
Robot-specific constraints, such as an unavailable hand,
are incorporated into the digital twin.

\begin{table*}[!t]
    \centering

    \caption{
        \textbf{HOI generation on ARCTIC.}
Baselines share the same split and first-frame conditioning; bimanual and whole-body metrics use retrained evaluators.
        $\dagger$: MAD-HOI is reimplemented and retrained;
EmbodyHOI applies inference-time guidance to our backbone.
        -- indicates unavailable metrics for hand-only methods;
        \textbf{Bold}/\underline{underline}: best/second best; $\rightarrow$GT: closer to ground truth is better.
    }

    \label{tab:exp_generation}

    \begingroup
    \renewcommand{\arraystretch}{1.14}
    \setlength{\tabcolsep}{3pt}
    \footnotesize

    \begin{tabular*}{\textwidth}{
        @{\extracolsep{\fill}}
        lcccccccccccc
        @{}
    }
        \toprule

        \multirow{2}{*}{\textbf{Method}}
        & \multicolumn{5}{c}{\textbf{Bimanual HOI}}
        & \multicolumn{4}{c}{\textbf{Whole-body HOI}}
        & \multicolumn{3}{c}{\textbf{Physical Quality}} \\

        \cmidrule(lr){2-6}
        \cmidrule(lr){7-10}
        \cmidrule(l){11-13}

        & FID $\downarrow$
        & R@3 $\uparrow$
        & MM-Dist $\downarrow$
        & Div $\rightarrow$ GT
        & Wrist (mm) $\downarrow$
        & FID $\downarrow$
        & R@3 $\uparrow$
        & MM-Dist $\downarrow$
        & AVE $\downarrow$
        & CR $\rightarrow$ GT
        & Pen (mm) $\downarrow$
        & FS (m/s) $\downarrow$ \\

        \midrule

        GT
        & 0.00 & 0.788 & 3.38 & 10.72 & 0.0
        & 0.00 & 0.720 & 3.79 & 0.0000
        & 0.100 & 5.9 & 0.0022 \\

        \midrule

        EmbodyHOI$\dagger$
        & 5.13 & 0.576 & \underline{5.24} & 9.64
        & \underline{77.9}
        & \underline{2.32} & \underline{0.629}
        & \underline{4.59} & \underline{0.0087}
        & 0.040 & 7.6 & \underline{0.0064} \\

        Text2HOI
        & \underline{1.51} & 0.523 & 5.54 & 9.68
        & 92.4
        & -- & -- & -- & --
        & 0.090 & -- & -- \\

        MAD-HOI$\dagger$
        & 5.63 & \underline{0.583} & 5.45
        & \underline{9.77} & 104.8
        & -- & -- & -- & --
        & 0.082 & -- & -- \\

        HandMDM
        & 25.59 & 0.308 & 8.35 & 9.10 & 161.3
        & -- & -- & -- & --
        & 0.022 & -- & -- \\

        HOIDiNi
        & 8.22 & 0.477 & 6.30 & 8.80 & 151.6
        & 23.65 & 0.311 & 8.14 & 0.0147
        & 0.168 & 11.9 & 0.0462 \\

        HOI-FHLI
        & 12.56 & 0.368 & 7.31 & 8.03 & 143.7
        & 37.65 & 0.265 & 9.47 & 0.0188
        & 0.079 & \textbf{4.6} & 0.0180 \\

        CoDA
        & 8.85 & 0.507 & 5.98 & 9.17 & 141.4
        & 71.20 & 0.244 & 10.72 & 0.0218
        & \textbf{0.103} & \underline{7.4} & 0.0433 \\

        \midrule

        \textbf{FineDex}
        & \textbf{0.65}
        & \textbf{0.701}
        & \textbf{4.02}
        & \textbf{10.58}
        & \textbf{58.3}
        & \textbf{1.61}
        & \textbf{0.704}
        & \textbf{3.83}
        & \textbf{0.0075}
        & \underline{0.092}
        & 7.8
        & \textbf{0.0030} \\

        \bottomrule
    \end{tabular*}
    \endgroup

\end{table*}

\paragraph{Interaction retargeting.}
We retarget generated hand motion with an interaction mesh over hand
and object keypoints in the object frame, preserving its Laplacian
coordinates under joint-limit and non-penetration constraints. The
resulting wrist and finger references guide the construction and
evaluation of contact programs; replaying them unchanged is only one
program option.

\paragraph{Contact programs.}
We partition the generated object articulation into $S$ opening or
closing segments. A contact program is represented by
\begin{equation}
    \pi=(m,\kappa_{1:S},b,\boldsymbol{\eta}),
    \label{eq:hoi_contact_program}
\end{equation}
where $m$ specifies the manipulating hand and object-motion plan,
$\kappa_s$ is a part-relative wrist pose and finger configuration
for segment $s$, $b$ denotes the supporting-hand behavior, and
$\boldsymbol{\eta}$ contains approach, pressing, timing, and closure
parameters. Candidate contacts include those extracted from the
generated interaction, geometrically synthesized push or hook
contacts, and human grasps from a library built from non-test ARCTIC
sequences.
Geometric candidates are ranked by their ability to drive the desired
hinge motion and their proximity to the generated contact. The HOI prior guides contact selection and support-hand placement
without fixing the final contact poses or execution timing.

Each program is compiled into wrist and finger targets
by smoothly blending free-space motion with object-relative
contact configurations initialized from the generated HOI.

\paragraph{Agent planning and digital-twin practice.}
We instantiate a multimodal planning agent with GPT-6 Astra
by default and Claude Fable 5.2 as an alternative
(Tab.~\ref{tab:exp_robot_ablations}).
Given the instruction, object state, rendered HOI keyframes,
and structured contact descriptors, the agent proposes
candidate contact programs and predicts their outcomes.
Each program is evaluated in parallel digital twins under
randomized dynamics.
The agent uses task outcomes, contact statistics, and
failure renders to diagnose failures and iteratively
refine its proposals.

Candidates are ranked by empirical success rate, with ties
broken by articulation progress, execution penalties,
and deviation from the generated HOI prior.
The top four goal-reaching candidates undergo additional
randomized rehearsals, and the most reliable program
$\pi^{\star}$ is selected for execution.
The committed program is executed open-loop without
online replanning.

We evaluate FineDex on floating Wuji hands and a Unitree G1
humanoid equipped with Wuji dexterous hands.
The selected program is executed using embodiment-specific
controllers.

\subsection{FingerBench}
\label{sec:method_fingerbench}

Conventional motion metrics do not assess whether
fine-grained instructions are followed.
FingerBench evaluates instruction compliance through two tracks.
The \emph{generation track} evaluates hand selection (Switch/Still),
finger motion (A1--A5), and finger contact (B1--B4).
A1--A5 test single-finger motion, same- and opposite-direction
finger motions, bimanual control, and action timing.
Success requires the instructed motion while preserving
uncommanded fingers.
B1--B4 evaluate contact with the reference finger,
exclusive contact, counterfactual finger contact,
and thumb--index pinching.
Success requires the designated contact and corresponding
object articulation, with unintended contacts excluded
where specified.

The \emph{robot track} evaluates physical execution using
finger-conditioned success rate (FC-SR) and Robo-SR.
Both require task completion and sustained contact by
the designated finger; FC-SR measures contact dominance
by impulse share, while Robo-SR limits the duration
of unintended contacts by other hand regions.

%% file: sec/4_experiment.tex
\section{Experiments}
\label{sec:experiments}

\begin{table*}[!t]
    \centering

\caption{\textbf{FingerBench results.} Success rates (\%).
A1--A5: single, same/opposite direction, bimanual, temporal; B1--B4: GT, exclusive, counterfactual, pinch.
Switch/Still: hand selection; robot evaluation uses floating Wuji hands and a shared search planner.}

    \label{tab:fingerbench}

    \begingroup
    \renewcommand{\arraystretch}{1.14}
    \setlength{\tabcolsep}{2.4pt}
    \footnotesize

    \begin{tabular*}{\textwidth}{
        @{\extracolsep{\fill}}
        l*{13}{c}
        @{}
    }
        \toprule

        \multirow{2}{*}{\textbf{Method}}
        
        & \multicolumn{5}{c}{\textbf{Finger Motion Control}}
        & \multicolumn{4}{c}{\textbf{Finger Contact Control}}
        & \multicolumn{2}{c}{\textbf{Hand Selection}}
        & \multicolumn{2}{c}{\textbf{Robotic Execution}} \\

        \cmidrule(lr){2-6}
        \cmidrule(lr){7-10}
        \cmidrule(lr){11-12}
        \cmidrule(l){13-14}

        & A1 & A2 & A3 & A4 & A5
        & B1 & B2 & B3 & B4
        & Switch & Still
        & FC-SR $\uparrow$ & Robo-SR $\uparrow$ \\

        \midrule

        GT
        & -- & -- & -- & -- & --
        & -- & -- & -- & --
        & -- & --
        & 6.2 & 5.4 \\

        \midrule

        EmbodyHOI$\dagger$
        & 0.0 & 0.0 & 0.0 & 0.0 & 0.0
        & \underline{16.7} & \underline{1.2} & 0.6 & 0.8
        & 0.0 & 0.0
        & 0.9 & 0.9 \\

        Text2HOI
        & 0.0 & 0.0 & 0.0 & 0.0 & 0.0
        & 15.5 & 0.8 & 0.4 & 0.5
        & 1.1 & 0.0
        & \underline{9.4} & 6.2 \\

        MAD-HOI$\dagger$
        & 0.0 & 0.0 & 0.0 & 0.0 & 0.0
        & 14.3 & 1.0 & 0.4 & 0.3
        & \underline{8.0} & 0.0
        & 1.3 & 0.9 \\

        HandMDM
        & 0.0 & 0.0 & 0.0 & 0.0 & 0.0
        & 0.0 & 0.0 & 0.0 & 0.0
        & 5.7 & 0.0
        & 7.6 & \underline{7.1} \\

        HOIDiNi
        & 0.0 & 0.0 & 0.0 & 0.0 & 0.0
        & 6.6 & \underline{1.2} & 0.4 & 0.3
        & 0.0 & 0.0
        & 2.7 & 0.0 \\

        HOI-FHLI
        & 0.0 & 0.0 & 0.0 & 0.0 & 0.0
        & 7.9 & \underline{1.2} & \underline{0.8} & \underline{1.1}
        & 0.0 & 0.0
        & 1.3 & 1.3 \\

        CoDA
        & 0.0 & 0.0 & 0.0 & 0.0 & 0.0
        & 2.7 & 0.2 & 0.2 & 0.0
        & 0.0 & 0.0
        & 2.7 & 3.1 \\

        \midrule

        \textbf{FineDex}
        & \textbf{92.5} & \textbf{91.2} & \textbf{37.0} & \textbf{73.4} & \textbf{45.6}
        & \textbf{55.8} & \textbf{60.1} & \textbf{64.9} & \textbf{57.4}
        & \textbf{83.9} & \textbf{100.0}
        & \textbf{21.9} & \textbf{19.2} \\

        \bottomrule
    \end{tabular*}
    \endgroup

\end{table*}

\subsection{Experimental Setup}
\label{sec:exp_setup}

\paragraph{Dataset.}
ARCTIC~\cite{ARCTIC} provides 3D body, hand, and object
motion for bimanual interactions with 11 articulated objects.
We split ARCTIC into training, validation, and test sets
and extract 32-frame windows with a stride of 16.
Each window is annotated with hand roles, finger-level contacts,
object-part regions, and articulation, first generated with
Gemini 3.8 and then verified with Claude, which corrected 497
of 13,472 annotations, mostly hand assignments and part references.

\paragraph{Metrics.}
We evaluate HOI generation using standard metrics:
FID, R-precision (R@3), matching distance (MM-Dist),
diversity (Div), wrist error, average variance error (AVE),
contact ratio (CR), penetration (Pen), and foot skating (FS).
To assess fine-grained controllability, FingerBench
(Sec.~\ref{sec:method_fingerbench}) evaluates hand selection,
finger motion, and designated contacts.
Robotic execution is evaluated using task success,
finger-conditioned success rate (FC-SR), and Robo-SR
under randomized dynamics.

\subsection{HOI Generation}
\label{sec:exp_generation}

\paragraph{Quantitative Comparison.}
For a fair comparison, we retrain all baselines on
the same ARCTIC split with our fine-grained annotations
and identical first-frame conditioning.
As shown in Tab.~\ref{tab:exp_generation}, FineDex
achieves the best results across all bimanual and
whole-body HOI metrics, reducing FID from
$1.51$ to $0.65$ for bimanual HOI and from
$2.32$ to $1.61$ for whole-body HOI.
It also achieves the lowest foot skating,
while maintaining diversity and contact ratio close to
ground truth.

\begin{figure}[!t]
    \centering
    \includegraphics[width=\columnwidth]{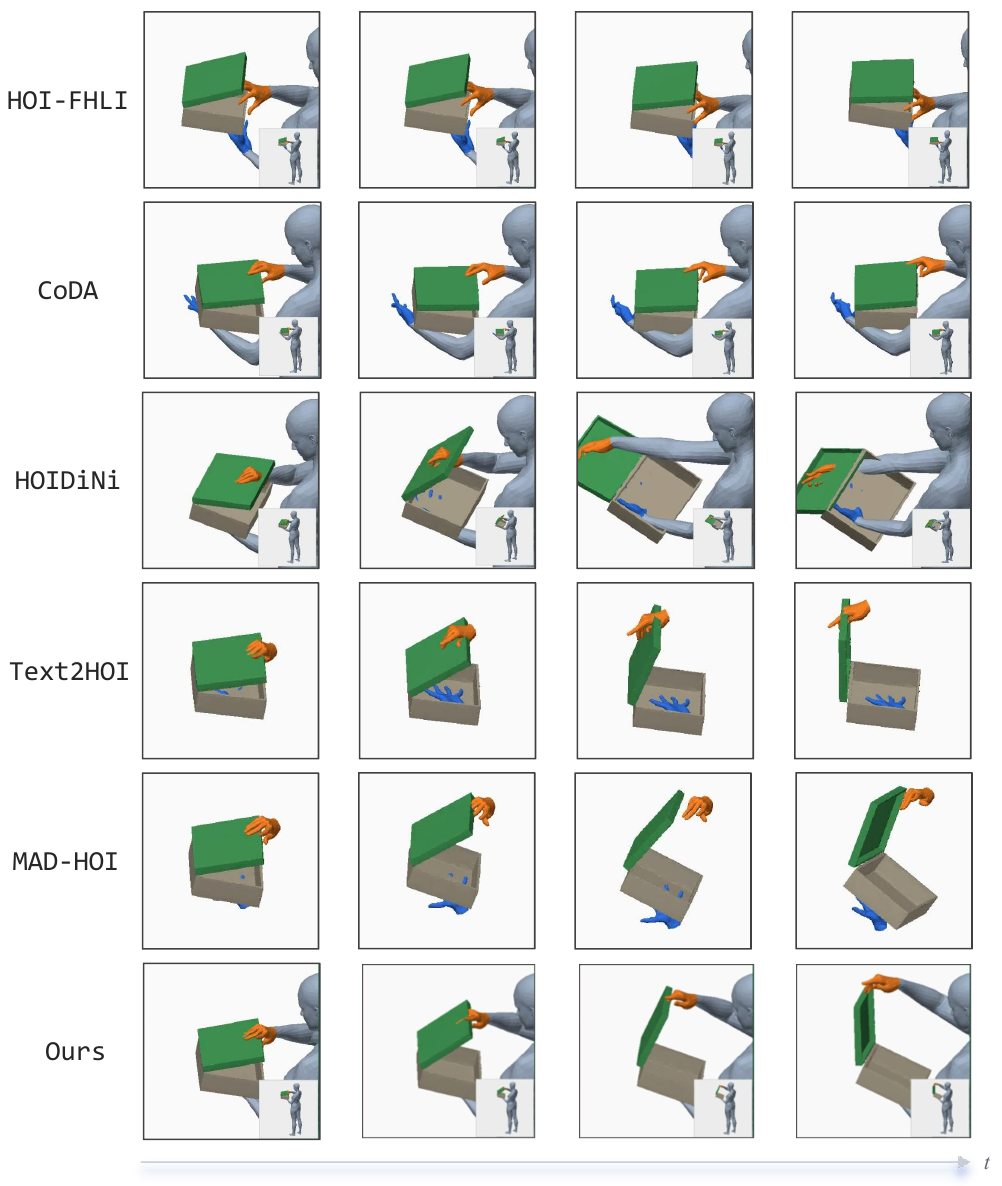}
    \caption{
        \textbf{Qualitative comparison on ARCTIC.}
        FineDex generates coordinated hand--object motion.
    }
    \label{fig:compare_baseline}
    \vspace{-2pt}
\end{figure}

\paragraph{Qualitative Comparison.}
Fig.~\ref{fig:compare_baseline} compares generated
HOI sequences on articulated objects.
FineDex produces coordinated hand--object motion
with sustained contact and substantial object articulation.
In contrast, HOI-FHLI\cite{HOIFHLI} and CoDA\cite{CoDA} exhibit limited articulation,
while HOIDiNi\cite{HOIDiNi} shows less consistent hand--object coordination
in the illustrated sequence.
MAD-HOI\cite{MADHOI} and Text2HOI\cite{Text2HOI} generate hand-only motion.

\begin{table*}[!t]
    \centering
    \caption{\textbf{Robot execution ablations.}
    (a) Effects of planning, practice, and HOI priors on task success.
    (b) Effects of finger instructions and instruction-aligned
    priors on finger-conditioned execution.
    All results are percentages; parentheses denote
    unconstrained task success.}
    \label{tab:exp_robot_ablations}

    \begingroup
    \renewcommand{\arraystretch}{1.13}
    \setlength{\tabcolsep}{4pt}
    \footnotesize

    \begin{minipage}[t]{0.52\textwidth}
        \centering

        \textbf{(a) Planning and motion priors}\\[4pt]

        \begin{tabular*}{\linewidth}{
            @{\extracolsep{\fill}}
            lcc
            @{}
        }
            \toprule
            \textbf{Method}
            & \textbf{Floating} $\uparrow$
            & \textbf{G1} $\uparrow$ \\
            \midrule

            Direct replay
            & 6.2 & 13.3 \\

            FineDex w/o practice
            & 19.5 & 11.7 \\

            Search planner + priors
            & 48.4 & 39.1 \\

            Astra w/o priors
            & 68.8 & 35.2 \\

            Astra w/o priors or library
            & 59.4 & 28.9 \\
            
            \midrule

            FineDex (GPT-6 Astra)
            & 75.0
            & 42.2 \\

            FineDex (Claude Fable 5.2)
            & 77.3
            & 46.9 \\

            \bottomrule
        \end{tabular*}
    \end{minipage}
    \hfill
    \begin{minipage}[t]{0.45\textwidth}
        \centering

        \textbf{(b) Finger-conditioned priors}\\[4pt]

        \begin{tabular*}{\linewidth}{
            @{\extracolsep{\fill}}
            lccc
            @{}
        }
            \toprule
            \textbf{Condition}
            & \textbf{Task} $\uparrow$
            & \textbf{FC-SR} $\uparrow$
            & \textbf{FC/Task} $\uparrow$ \\
            \midrule

            Astra
            & (78.1) & 6.7 & 8.6 \\

            \midrule

            Astra + finger instr.
            & 40.2 & 19.4 & 48.2 \\

            \quad + unedited prior
            & 29.7 & 15.2 & 51.1 \\

            \quad + finger-edited prior
            & \textbf{40.3} & \textbf{24.4}
            & \textbf{60.5} \\

            \bottomrule
        \end{tabular*}
    \end{minipage}

    \endgroup
\end{table*}

\begin{table}[!t]
    \centering
    \caption{\textbf{Object configuration generalization.}
        Task success (\%) under changes in placement, size, and initial articulation of seen objects.
        W/o priors removes generated HOI while retaining the grasp library and geometric contacts.}
    \label{tab:exp_robot_robustness}
    \begingroup
    \renewcommand{\arraystretch}{1.1}
    \setlength{\tabcolsep}{3pt}
    \footnotesize
    \begin{tabular*}{\columnwidth}{@{\extracolsep{\fill}}llcccc@{}}
        \toprule
        \textbf{Robot} & \textbf{Method} & \textbf{Placement} & \textbf{Size} & \textbf{Angle} & \textbf{All} \\
        \midrule
        \multirow{2}{*}{Floating} & w/o priors & 65.2 & 59.6 & 68.8 & 63.4 \\
         & FineDex & \textbf{69.1} & \textbf{68.9} & \textbf{75.8} & \textbf{70.3} \\
        \midrule
        \multirow{2}{*}{G1} & w/o priors & 38.3 & 29.8 & 50.0 & 36.6 \\
         & FineDex & \textbf{39.1} & \textbf{32.1} & \textbf{57.8} & \textbf{39.4} \\
        \bottomrule
    \end{tabular*}
    \endgroup
\end{table}

\subsection{FingerBench Evaluation}
\label{sec:exp_finger}

\begin{figure}[!t]
    \centering
    \includegraphics[width=\linewidth]{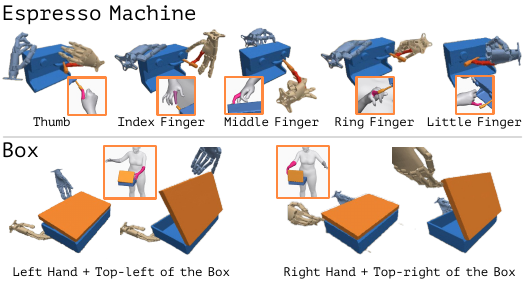}
    \caption{
        \textbf{Qualitative comparison of fine-grained control.}
        Top: finger-specific manipulation using different
        instructed fingers. Bottom: hand selection and designated contact
        control of an articulated object.
    }
    \label{fig:control}
\end{figure}

As shown in Tab.~\ref{tab:fingerbench}, we evaluate FingerBench on nine shared ARCTIC objects.
Fig.~\ref{fig:control} illustrates finger-specific manipulation
and instruction-guided control over hand selection
and contact locations.

\paragraph{Fine-Grained Controllability and Execution.}
On FingerBench, FineDex achieves
$83.9\%$ Switch and $100.0\%$ Still success in hand selection,
while no baseline exceeds $8.0\%$ Switch.
Existing HOI generators lack explicit finger-level
control mechanisms and score $0\%$ on A1--A5,
even when given the same instructions as text.
FineDex achieves $37.0$--$92.5\%$ across these tasks.
On counterfactual finger contact, FineDex reaches $64.9\%$
versus $0.8\%$ for the strongest baseline.
In physics-based execution, FineDex achieves $21.9\%$ FC-SR
and $19.2\%$ Robo-SR, more than doubling the strongest baseline.
Using captured human motion as the prior yields only
$6.2\%$ FC-SR, as its finger contacts do not necessarily
match the instructions.

\paragraph{Finger Control Ablations.}
Tab.~\ref{tab:finger_ablation} evaluates the effects of
dual-trajectory sampling and finger-specific contact refinement.
For finger motion, single-trajectory conditioning causes
$11.9^\circ$ of unintended finger motion and reduces
single-finger success from $92.5\%$ to $41.2\%$.
In contrast, our masked dual-trajectory sampling preserves
uncommanded fingers by construction.
Even with our finger-edit training data, HandMDM\cite{handmdm} achieves
$65.6\%$ finger identification but only $3.8\%$ success,
as other fingers move with the commanded one.
For finger contact, fingertip attraction alone performs
well on B1 but fails under exclusive contact constraints
(B2--B4), with unintended contacts in $97.9\%$ of trials.
Our full objective reduces this rate to $41.5\%$
and substantially improves counterfactual contact success (B3)
from $0.8\%$ to $64.9\%$.

\begin{table}[!t]
    \centering
        \caption{
        \textbf{Ablation of finger-level control.}
        Top: single- vs.\ dual-trajectory sampling for finger motion;
        Leak.\ measures changes in uncommanded fingers.
        Bottom: contact refinement with fingertip attraction
        (Contact IK), penetration avoidance (+pen.),
        and the full objective.
        Other denotes unintended contact rate (\%).
    }
    \label{tab:finger_ablation}
    \begingroup
    \renewcommand{\arraystretch}{1.1}
    \setlength{\tabcolsep}{3pt}
    \footnotesize
    \begin{tabular*}{\columnwidth}{@{\extracolsep{\fill}}lcccccc@{}}
        \toprule
        \textbf{Finger motion} & A1 & A2 & A3 & A4 & A5 & Leak.\,($^\circ$)\,$\downarrow$ \\
        \midrule
        Single trajectory
            & 41.2 & 13.8 & 3.7 & 2.2 & 6.4 & 11.9 \\
        Dual trajectory (ours)
            & \textbf{92.5} & \textbf{91.2} & \textbf{37.0} & \textbf{73.4} & \textbf{45.6} & \textbf{0.0} \\
        \midrule
        \textbf{Finger contact} & B1 & B2 & B3 & B4 & Other\,$\downarrow$ & \\
        \midrule
        Contact IK
            & 50.8 & 1.6 & 0.8 & 1.4 & 97.9 & \\
        Contact IK\,+\,pen.
            & \textbf{56.0} & 1.0 & 1.0 & 1.1 & 98.2 & \\
        Full objective (ours)
            & 55.8 & \textbf{60.1} & \textbf{64.9} & \textbf{57.4} & \textbf{41.5} & \\
        \bottomrule
    \end{tabular*}
    \endgroup
\end{table}

\subsection{Robot Execution}
\label{sec:exp_robot}

\begin{figure*}[!t]
    \centering
    \includegraphics[width=\textwidth]{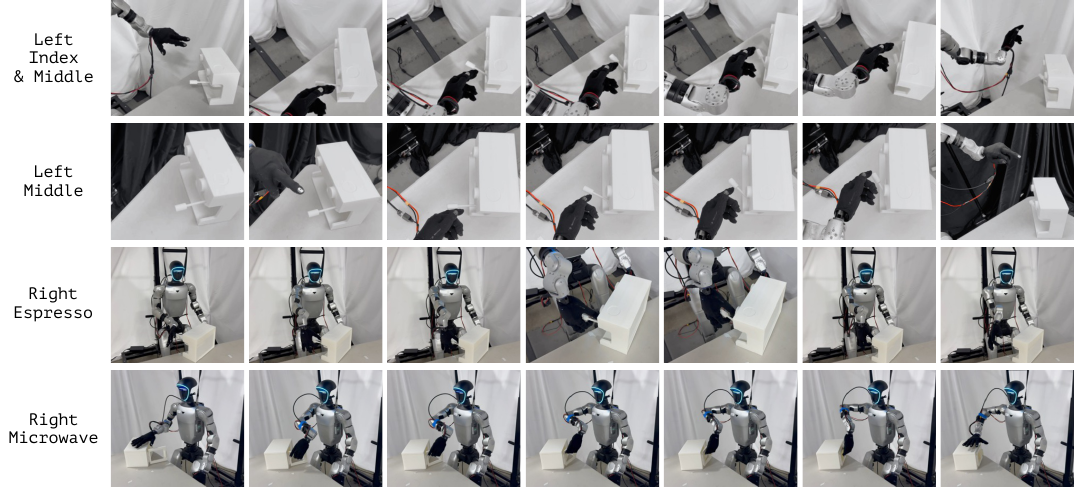}
    \caption{
        \textbf{Real-world humanoid robot dexterous manipulation.}
        G1 humanoid manipulation with different hand and
        finger instructions across articulated objects.
    }
    \label{fig:real_robot}
\end{figure*}

\paragraph{Object Configuration Generalization.}
As shown in Tab.~\ref{tab:exp_robot_robustness},
FineDex improves task success over Astra-only planning
in every condition. The gains are consistent but modest:
across 174 shifted configurations, FineDex achieves
$54.8\%$ task success, compared with $50.0\%$
without generated priors, with the same planning budget and no
retraining. The largest gain with floating hands occurs under
object-size variations ($68.9\%$ vs.\ $59.6\%$).

\paragraph{Role of the HOI Prior.}
The generated HOI guides contact selection rather than
prescribing a trajectory to replay.
Only $21\%$ of committed programs directly reuse
the generated contacts; more often, the prior guides
the ranking of library and synthesized contacts
and the placement of the support hand.
Among 14 configurations that succeed in at least half
of the trials with priors but never without them,
10 use the generated support-hand pose to stabilize
the object, a strategy not found without the prior. Removing the grasp library as well leaves no human interaction data and lowers success to $59.4\%$ (floating) and $28.9\%$ (G1) in
Tab.~\ref{tab:exp_robot_ablations}(a); no box trial then succeeds, since
geometric synthesis finds no contact that opens the lid.

The importance of instruction-aligned priors is further
evident in Tab.~\ref{tab:exp_robot_ablations}(b).
With finger instructions alone, the agent achieves
$40.2\%$ task success and $19.4\%$ FC-SR.
Adding an unedited prior reduces these to $29.7\%$
and $15.2\%$, respectively, as its suggested contacts
may conflict with the instructed finger.
In contrast, finger-edited priors restore task success
to $40.3\%$ and improve FC-SR to $24.4\%$,
with $60.5\%$ of successful trials using the
designated finger. These results highlight the importance of preserving
finger-level intent during robotic execution.

\paragraph{Planning and Practice.}
In Tab.~\ref{tab:exp_robot_ablations}(a), directly replaying
the retargeted generated motion achieves only $6.2\%$ task
success, compared with $75.0\%$ after practice.
The agent alone does not close this gap: executing its most
confident first-round program without rehearsal succeeds in
$19.5\%$ of floating-hand and $11.7\%$ of G1 trials.
Success grows with the practice budget: stopping after one or
two of the four rounds yields $43.8\%$ and $57.0\%$ with floating
hands ($25.0\%$ and $30.5\%$ on G1). Gains level off after about six.
Claude Fable 5.2 performs comparably to Astra without
model-specific tuning.

\paragraph{Finger-constrained execution.}
Tab.~\ref{tab:exp_robot_ablations}(b) evaluates finger-specific
instructions provided only at test time.
Stating the finger raises FC-SR from $6.7\%$ to $19.4\%$ without a generated
prior. A prior that ignores the instruction lowers task success to $29.7\%$;
editing it to the designated finger restores task success ($40.3\%$) and gives
the highest FC-SR ($24.4\%$) and FC/Task ($60.5\%$).


\paragraph{Real-world dexterous manipulation.}
Fig.~\ref{fig:real_robot} demonstrates FineDex on a physical
Unitree G1 humanoid equipped with Wuji Hand 2.2,
covering finger-specific left-hand manipulation and
right-hand operation of an espresso machine and a microwave.
The contact programs selected through digital-twin practice
are transferred to the robot and executed open-loop
without online replanning.
To align the physical setup with simulation,
the digital twin selects a robot stance $\mathbf{T}^{\mathcal{P}}$
that maximizes the fraction of reachable, collision-free
wrist poses. The object is then placed at
$(\mathbf{T}^{\mathcal{P}})^{-1}\mathbf{T}^{\mathcal{O}}_1$
in the pelvis frame, without further robot-side tuning.
Under a fixed object placement, the left-middle-finger
program succeeds in all five consecutive trials.
The digital twin uses the printed objects' meshes and masses,
with control gains fixed across tasks.

%% file: sec/5_conclusion.tex
\section{Conclusion}
\label{sec:conclusion}

We presented FineDex, a framework that makes finger-level intent
both expressible and executable. Trained on our fine-grained annotations, the generator produces coordinated human--object motion conditioned on hand, finger, and contact instructions. Rather than directly replaying generated motions,
an agent uses them as priors to discover executable contact
behaviors through test-time practice. Experiments demonstrate
state-of-the-art fine-grained controllability and HOI generation
quality, substantial gains in dexterous manipulation success,
and successful execution on a real humanoid robot.
We further introduce FingerBench to evaluate instruction
following in both generation and physical execution.

\paragraph{Limitations.}
Enforcing designated-finger constraints reduces task success,
while limited arm reach and imperfect wrist tracking remain
major failure modes on the humanoid.
Real-robot behaviors are executed open loop, without
online feedback; closing this loop is left to future work.


